\documentclass[letterpaper, 10 pt, conference]{ieeeconf}  % Comment this line out if you need a4paper

\IEEEoverridecommandlockouts                              % This command is only needed if 
\usepackage{tikz}
\usepackage{lipsum}
\newcommand\copyrighttext{%
	\footnotesize \copyright 2022 IEEE. Personal use of this material is permitted. Permission from IEEE must be obtained for all other uses, in any current or future media, including reprinting/republishing this material for advertising or promotional purposes,creating new collective works, for resale or redistribution to servers or lists, or reuse of any copyrighted component of this work in other works}
\newcommand\copyrightnotice{%
	\begin{tikzpicture}[remember picture,overlay]
	\node[anchor=south,yshift=10pt] at (current page.south) {\fbox{\parbox{\dimexpr\textwidth-\fboxsep-\fboxrule\relax}{\copyrighttext}}};
	\end{tikzpicture}%
}
\usepackage{cite}
\usepackage{amsmath,amssymb,amsfonts}
\usepackage{algorithmic}
\usepackage{graphicx}
\usepackage{textcomp}
\usepackage{xcolor}
\usepackage{amssymb}
\usepackage{fixmath}
\usepackage{comment}
\usepackage{balance}
\usepackage{multicol}
\usepackage{url}
\usepackage{gensymb}
\usepackage[T1]{fontenc}
\usepackage{float}
\usepackage{mwe}
\usepackage{subfig}
\DeclareMathSizes{10}{10}{3}{3}

\def\BibTeX{{\rm B\kern-.05em{\sc i\kern-.025em b}\kern-.08em
		T\kern-.1667em\lower.7ex\hbox{E}\kern-.125emX}}

\newcommand{\rvect}[1]{\begin{bmatrix} #1 \end{bmatrix}} %row vector
\newcommand{\cvect}[1]{\begin{bmatrix}#1\end{bmatrix}}

\title{\LARGE \bf
Recovering the 3D UUV Position using UAV Imagery in Shallow-Water Environments
}

\author{Antun \DJ ura\v{s}$^{1}$, Matija Sukno and Ivana Palunko% <-this % stops a space
	\thanks{*This work is supported in part by project SeaClear, European Union's Horizon 2020 research and innovation programme under grant agreement No. 871295, in part by Croatian Science Foundation under the project DOK-2020-01-8228 and in part by project InnovaMare, Interreg IT-HR, European Regional and Development Fund under No.10248782.}% <-this % stops a space
	\thanks{$^{1}$All authors are with Laboratory for Intelligent Autonomous Systems (LARIAT), Department of Electrical Engineering and Computing, University of Dubrovnik, Croatia {\tt\small antun.djuras@unidu.hr};{\tt\small matija.sukno@unidu.hr};
		{\tt\small ivana.palunko@unidu.hr}}%
}

\begin{document}

\maketitle
\thispagestyle{empty}
\pagestyle{empty}
\copyrightnotice
\begin{abstract}
	In this paper we propose a novel approach aimed at recovering the 3D position of an UUV from UAV imagery in shallow-water environments. Through combination of UAV and UUV measurements, we show that our method can be utilized as an accurate and cost-effective alternative when compared to acoustic sensing methods, typically required to obtain ground truth information in underwater localization problems. Furthermore, our approach allows for a seamless conversion to geo-referenced coordinates which can be utilized for navigation purposes. To validate our method, we present the results with data collected through a simulation environment and field experiments, demonstrating the ability to successfully recover the UUV position with sub-meter accuracy.
\end{abstract}

%%%%%%%%%%%%%%%%%%%%%%%%%%%%%%%%%%%%%%%%%%%%%%%%%%%%%%%%%%%%%%%%%%%%%%%%%%%%%%%%
\section{Introduction}
Unavailability of Global Positioning System (GPS) information underwater makes the task of Unmanned Underwater Vehicle (UUV) localization a difficult problem that requires deployment of expensive acoustic sensors such as Doppler Velocity Log (DVL), Long BaseLine (LBL) and Ultra-Short BaseLine (USBL), typically fused with Inertial Measurements Units (IMUs). Sensor fusion of cost effective alternatives based on visual information obtained via cameras or imaging sonars with typical sensor suite (IMUs, pressure sensors, range finders) for pose-estimation is proposed in \cite{Guth2014} but lacks robustness and accuracy of standard approaches. One of the major problems hindering development of underwater localization and object tracking methods, compared to their terrestrial counterparts, is lack of cost-effective solutions for obtaining the ground truth. These factors result in scarcity of benchmark datasets crucial for assessing the quality of the developed pose estimation methods. 
\begin{figure}[!htb]
	\centering
	\includegraphics[width=0.90\columnwidth]{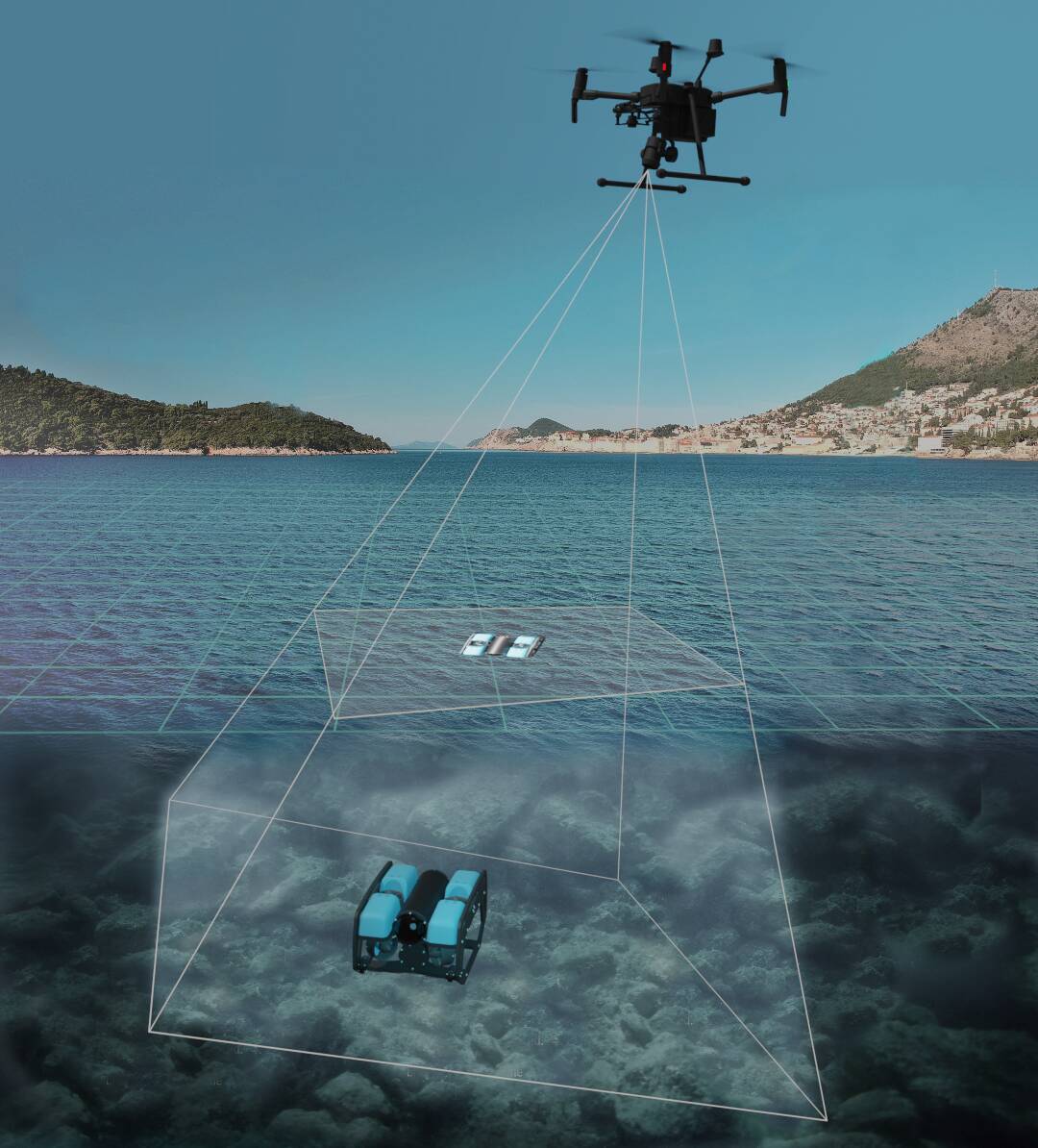}
	\caption{Illustration of the main components of our method, the UAV monitoring the operating area of the UUV} 
	\label{fig:main-image}
\end{figure}

This paper proposes an alternative approach to ground truth acquisition for recovering movement in the \textit{x-y} plane, suffering from unbounded drift when dead reckoning techniques are employed to obtain position estimates \cite{Loebis2002}. Here we combine the information from an Unmanned Aerial vehicle (UAV), equipped with a GPS and a monocular camera, surveying a shallow-water area in which UUV operates and depth readings from the pressure sensor equipped to the UUV. This allows us to resolve the depth ambiguity inherent to a monocular system and localize the UUV in terms of its 3D position with respect to the UAV camera. Furthermore, it allows us to position the UUV in terms of GPS coordinates, thus making a cost-effective solution for creating geo-referenced annotated datasets in shallow-water environments. We first demonstrate our approach on simulated data obtained through ROS (Robot Operating System) \cite{Quigley09} and Gazebo Simulator \cite{GazeboKoenig}, to confirm the methodology in presence of minimal sources of noise. After that results for experiments performed on real systems are presented and compared to tape-measured ground truth.

Main contributions of this paper are:
\begin{itemize}
	\item We utilize UAV altitude and UUV depth readings to recover the 3D position of the UUV, from 2D pixel coordinates of the UUV tracked in the image sequences obtained by the UAV surveying the operating area.
	\item We recover the GPS position of the UUV, providing a cost effective solution for acquisition of geo-referenced datasets for underwater motion estimation in shallow-water environments.
	\item Proposed method is verified both in simulation and multiple field experiments.
\end{itemize}

The rest of this paper is structured as follows: Section~II gives an overview of the problem and main components of our method. Section III contains a brief summary of related work. In Section IV concepts relevant to the proposed method are briefly reviewed, followed by a rundown of our method. In Section V technical details of our data acquisition system are described, including the setup used to obtain the field experiment data, our approach to result validation and analysis of results. Finally in Section VI we present a conclusion and propose further research directions.

\section{Problem statement}

We consider a UUV operating in shallow-water environment with an UAV hovering above the operating area capturing imagery, as depicted in Figure \ref{fig:main-image}. Recovering the 3D position of the UUV is a 2 step process:
\begin{enumerate}
	\item Tracking the 2D image position of the UUV in the image sequence.
	\item Relating the 2D image coordinates to 3D coordinates in the coordinate system of the camera.
\end{enumerate}

Tracking of the UUV image position can be formulated as a computer vision problem of Single Object Tracking (SOT). Specifics of the light-water interaction introduce additional disturbances resulting in need for a robust tracking algorithm. Due to underwater light attenuation, visibility of the UUV decreases when operating at higher depths. Furthermore, due to water refraction at the surface 2D projection of the UUV captured by the UAV camera might undergo changes of object shape or occlusions. Recovering the 3D position of the UUV from its 2D projection is an inverse perspective projection, which requires knowledge of distance from the cameras optical center. In our method we utilize combined measurements of the UAV altitude and UUV depth to obtain this information. 

\begin{figure*}[!htb]
	\centering
	\includegraphics[width=\linewidth]{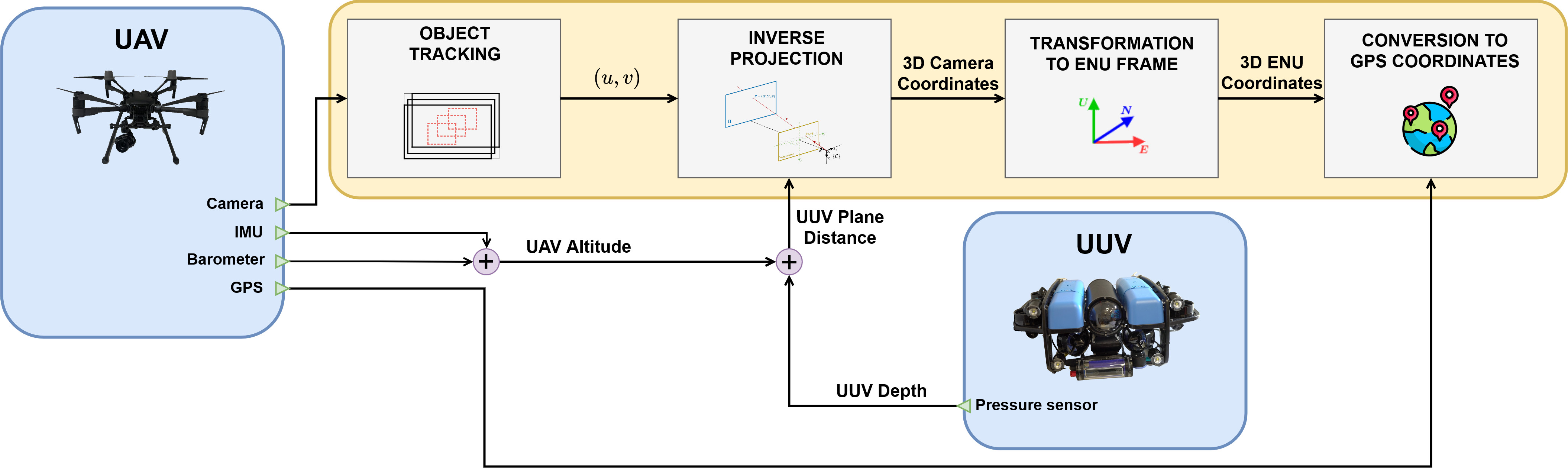}
	\caption{System diagram illustrating sensor information flow from the robots (blue) to main processing components (yellow) used to obtain the UUV position estimate.}
	\label{fig:system-scheme}
\end{figure*}

Finally, we utilize the GPS position of the UAV and 3D position of the UUV to recover the GPS coordinates of the UUV, enabling data acquisition of geo-referenced data. An overview of our approach, with regards to individual processing components and information flow from the robot sensors, is depicted in Figure \ref{fig:system-scheme}.

\section{Related Work}

Datasets from field experiments annotated with ground truth positions of an UUV are scarce. In literature, work involving evaluation of pose estimation methods in underwater environment mostly fall back to qualitative evaluation of trajectory tracking \cite{QuattriniLiISER2016} or employ alternative methods to avoid the use of expensive acoustic positioning systems. In \cite{Ferrera2018aqualoc} ground truth information is produced using the image based 3D reconstruction procedure (i.e. full batch off-line bundle adjustment). Mallios et. al \cite{Mallios2017} give tape-measured relative distances between sparse landmarks positioned along the trajectory as a source of ground truth for horizontal movement. Authors of \cite{HAUDSong2021} produced the first publicly available dataset with millimeter precision using the motion capture (MOCAP) system in an indoor pool.  

\begin{figure}
	\noindent \includegraphics[width=\columnwidth]{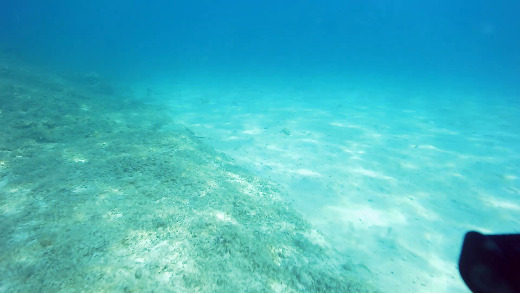}
	\caption{Image demonstrating the flickering effects of sunlight, often seen in shallow-water images captured by an UUV camera}
	\label{fig:shallow-water}
\end{figure}

Another approach that mitigates the need for measuring the ground truth data is using a simulation environment to obtain underwater datasets, since ground truth pose information is readily available. Simulation software such as UUV Sim \cite{Manhaes2016}  or UWSim \cite{Prats2012} focused on underwater domain can be used to simulate the interaction of the robot with the environment, implement control and navigation algorithms, and simulate auxiliary sensors. Simulating physically realistic underwater imagery is a difficult task that depends on modeling complex light-water interaction effects, which introduce both photometric and geometric distortions. Some recent advances focused on bridging the gap between real and synthetic imagery in underwater scenarios \cite{Zwilgmeyer2021ICCV} \cite{AlvarezTun2019} have been achieved, however producing realistic synthetic datasets for vision-based pose estimation remains an open problem. Shallow-water imagery is particularly difficult to simulate due to additional dependency on ambient light, manifesting as flickering caused by refraction of sunlight at water surface, as seen in Figure \ref{fig:shallow-water}. 

Single object tracking is a well studied problem in computer vision literature, focused on estimating the image position of the selected object in subsequent frames. Initial position of the object in the image is given by a rectangular bounding box. Typical disturbances that occur are occlusion, variations in illumination, change of viewpoint, rotation and motion blur. Current state-of-the-art algorithms can be divided into correlation-based filtering and deep-learning methods, while earlier work was focused on methods based on optical-flow and classic estimation algorithms such as Kalman Filter \cite{Zhang2021}. In our work we employ trackers available as part of the OpenCV library due to their versatility and simplicity of implementation. Evaluation of their properties and performance is conducted in \cite{Brdjanin2020}.

\section{Methodology}
This section briefly reviews relevant concepts for recovering the 3D and GPS positions of the UUV and provides a complete rundown of transformations used in our approach.
\subsection{The pinhole camera model}
\begin{figure}[!htb]
	\centering
	\includegraphics[width=0.9\columnwidth]{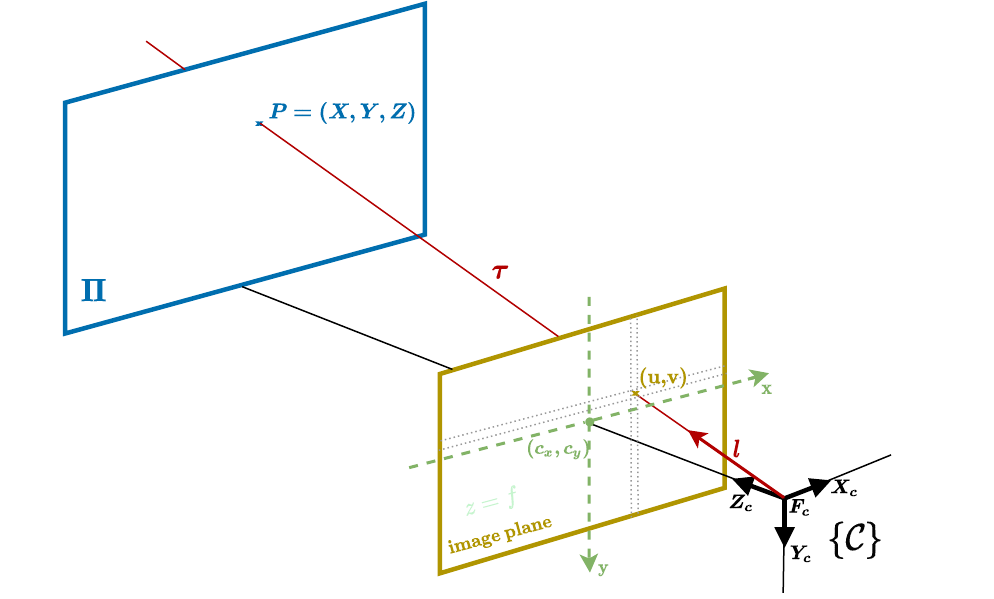}
	\caption{The pinhole camera model}
	\label{figPinholeModel}
\end{figure}

The pinhole camera model is the most commonly used camera model, which describes mathematical relationship between the points in 3-D space and points projected to 2-D image plane (Figure \ref{figPinholeModel}). This $\mathbb{R}^3\rightarrow\mathbb{R}^2$ mapping from $(X,Y,Z)$ point of the scene to point on the image plane is referred to as the perspective transformation, which is equivalent to the following (when $Z \ne 0$ ):
\begin{equation}
x' = X / Z \nonumber
\end{equation}
\begin{equation}
y' = Y / Z \nonumber
\end{equation}
where $(x',y')$ are normalized camera coordinates. Transformation between normalized coordinates to pixel coordinates is given by:
\begin{equation}
u  = f_{x} * x' + c_{x} \nonumber
\end{equation}
\begin{equation}
v  = f_{y} * y' + c_{y} \nonumber
\end{equation}
These expressions can be written in the following linear form:
\begin{equation}
\begin{bmatrix}
f_{x} & 0 & c_{x}\\
0 & f_{y} & c_{y}\\
0 & 0 & 1
\end{bmatrix}
\cvect{x' \\ y' \\ 1} = K \cvect{x' \\ y' \\ 1} = \cvect{u \\ v \\ 1}
\label{eq:normalized-to-px-coord}
\end{equation}

In the above formulation, $f_{x}$ and $f_{y}$ are focal lengths in pixel units while $(c_{x}, c_{y})$ is a principal point, which is usually at the image center. Matrix $K$ is typically referred to as the camera calibration matrix, where $f_{x}, f_{y}, c_{x}$ and $c_{y}$ are intrinsic parameters of the camera. 

The pinhole model makes the assumption aperture is a single point. However, in real world scenarios, cameras use lenses that introduce additional distortion \cite{Sturm2010}. For our application we consider a standard radial and tangential distortion model, extending the above pinhole model by adding the following polynomial terms:
\begin{equation}
x'' = x' + \sum_{i=1}^{3}{k_i r^{2i}} + 2p_{1}x'y' + p_{2}(r^2+2x'^2) \label{eq:dist-x}\\
\end{equation}
\begin{equation}
y'' = y' + \sum_{i=1}^{3}{k_i r^{2i}} + 2p_{1}x'y' + p_{2}(r^2+2y'^2) \label{eq:dist-y}\\
\end{equation}
where $r^2 = x'^2 + y'^2$ is the distance from image center, $[k_1,k_2,k_3]$ are radial distortion coefficients and $[p_{1}, p_{2}]$ are tangential distortion coefficients.
\subsection{Line-Plane Intersection}
Vector parametrized line equation is given as
\begin{equation}
p = l_{0} + ld \label{eq:line}
\end{equation}
where $l_{0}$ is a point on the line, $l$ is a vector in the direction of the line and $d \in \mathbb{R}$ is a scalar factor. Assuming a known plane
$$\Pi: (p-p_{0}) \cdot n = 0,$$ 
where $p_0$ is a point on the plane and $n$ is a plane normal, a point of the intersection can be calculated by substituting the line into the plane equation $\Pi$ and solving for $d$:
\begin{equation}
d = \frac{(p_{0} - l_{0})\cdot n}{l \cdot n}\\
\label{eq:solve-plane-vec}
\end{equation}

\subsection{Recovering the 3D UUV position}
To obtain the 2D pixel coordinates $(u,v)$ of the UUV from the image sequence captured by the UAV, we employ a OpenCV implementation of the Multiple Instance Learning (MIL) tracker. Our approach integrates altitude measurements $a_{UAV}$ of the UAV with depth measurements $d_{UUV}$ from the UUV to resolve the scene depth ambiguity of a single-view and find the mapping from 2D image coordinates to 3D camera coordinates. Assuming a known camera calibration matrix $K$ and distortion model parameters, 2D pixel coordinates of the UUV are transformed to normalized camera coordinates $(x', y')$ by inverting the expressions (\ref{eq:normalized-to-px-coord}) - (\ref{eq:dist-y})

\begin{equation}
(u,v) \overset{K^{-1}}{\longrightarrow} (x'',y'') \overset{undistort}{\longrightarrow} (x',y') \nonumber .
\end{equation}

By defining a line through optical center $F_{c}$, we can set $l_{0} = (0,0,0)$ and define a vector ${}^C l$ corresponding to normalized camera coordinates

\begin{equation}
{}^C l = (\frac{X}{Z}, \frac{Y}{Z}, 1) = (x', y', 1). \label{eq:l}  \nonumber
\end{equation}

Plane is parametrized by a point ${}^G p_0$ and a normal ${}^G n$ in the ENU coordinate frame $\{G\}$ with origin positioned at the UAV camera  
$$
a_{cam} = a_{UAV} - c_{offset, z}
$$
$$
{}^G p_0 = a_{cam} + d_{UUV}
$$
$$
{}^G n = \rvect{0 & 0 & 1}^T
$$
$c_{offset, z}$ is a vertical offset of a downward facing camera mounted on the UAV.
Points expressed in $\{G\}$ are related to camera coordinate frame $\{C\}$ by a rotation $R^C_G \in SO(3)$ obtained from UAV camera gimbal angle measurements:
\begin{equation}
R^G_{C'} = R_{z}(\Psi)R_{y}(\Theta)R_{x}(\Phi) \nonumber
\end{equation}
\begin{equation}
R^C_G = R^C_{C'}R^{C'}_{G} = R^C_{C'}(R^{G}_{C'})^T
\label{eq:gimbal-to-cam}
\end{equation}
where $(\Psi,\Theta,\Phi)$ are \textit{yaw-pitch-roll} angles representing the orientation of the camera with regards to $\{G\}$ and  $\{C'\}$ is an intermediate frame with the X axis pointing along the camera optical axis. To abide the Z-forward convention we perform an additional rotation $R^C_{C'}$. $R_{z}(\Psi),R_{y}(\Theta),R_{x}(\Phi)$ are elementary rotation matrices following passive convention.

By intersecting the plane $${}^C n = R^C_G {}^G n $$ $${}^C p_0 = R^C_G {}^G p_0$$ with the line defined by the vector ${}^C l$ based on equation (\ref{eq:solve-plane-vec}) a 3D position ${}^C P$ of the UUV is obtained in the camera coordinate frame as
\begin{equation}
d = \frac{{}^C p_{0} \cdot {{}^C n}}{{}^C l \cdot {{}^C n}}, \nonumber\\
\end{equation}
\begin{equation}
{}^C P = {}^C l d \nonumber\\
\end{equation}

\subsection{Conversion to GPS coordinates}
Estimated 3D positions of the UUV in the camera coordinate frame can be converted to geodetic coordinates to obtain geo-referenced data. UUV position ${}^C P $ is transformed back into the ENU coordinate frame $\{D\}$ fixed to the UAV body:
\begin{equation}
{}^D c_{offset} = R_{z}(\Psi_B)R_{y}(\Theta_B)R_{x}(\Phi_B) c_{offset} \nonumber
\end{equation}
\begin{equation}
{}^D P = R^G_C {}^C P + {}^D c_{offset} \nonumber
\end{equation}
where $(\Psi_B,\Theta_B,\Phi_B)$ are \textit{yaw-pitch-roll} angles representing orientation of the UAV body frame with respect to {D}, $c_{offset}$ is the camera position in the body frame of the UAV and $R^G_C$ is an inverse of equation \ref{eq:gimbal-to-cam}.

Conversion from local ENU coordinates ${}^D P$ to geodetic coordinates can be divided into 2 steps:
\begin{enumerate}
	\item Convert local ENU coordinates to Earth-Centered, Earth-Fixed (ECEF) coordinates:
	\item Convert ECEF coordinates to geodetic coordinates
\end{enumerate}

The Earth-Centered, Earth-Fixed (ECEF) is a right-handed Cartesian coordinate system  centered at the Earth's origin \cite{Fossen2011}. The +X axis passes through the Equator and Prime Meridian intersection, and the Z axis passes through the north pole. To transform the point in the local ENU frame a reference geodetic coordinate $(\phi, \lambda, h)$ of the origin is required. First the translation of the ENU origin in ECEF coordinates is determined:
\begin{align*} 
X_0 &= (N+h) \cos \phi \cos \lambda \\
Y_0 &= (N+h) \cos \phi \sin \lambda \\
Z_0 &= (\frac{r^2_p}{r^2_e} N + h)sin \phi
\end{align*}
where N is the distance from the surface to the Z-axis along the ellipsoid normal
$$
N = \frac{r_e^2}{\sqrt{r^2_e\cos^2(\phi)+r^2_p\sin^2(\phi)}}
$$
and $r_e$ and $r_p$,  equatorial and polar earth radii respectively, are the parameters of the reference ellipsoid. Conversion from local ENU coordinates to ECEF coordinates can then be expressed as:
\begin{equation}
{}^{ECEF} P = \begin{bmatrix}
-\sin\lambda & -\sin\phi\cos\lambda & \cos\phi\cos\lambda\\
\cos\lambda & -\sin\phi\sin\lambda & \cos\phi\sin\lambda\\
0 & \cos\phi & \sin\phi
\end{bmatrix}
{}^D P + \cvect{X_0 \\ Y_0 \\ Z_0} \nonumber
\end{equation}

ECEF positions of the UUV can be transformed to geodetic coordinates based on the procedure established by Heikkinen and cited in \cite{Zhu1994}. We utilize the open-source library \cite{MichGPS} which implements this solution to perform the conversion and will leave out the details of implementation for sparsity.

\section{Results}

\begin{figure}[t]
	\centering
	\includegraphics[width=0.85\columnwidth]{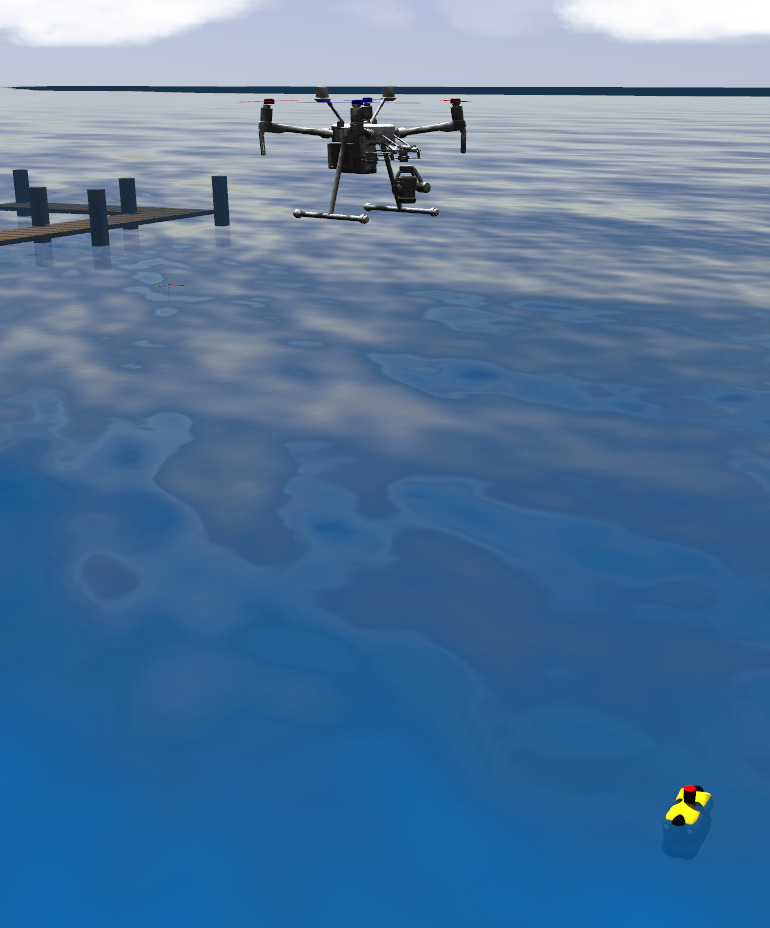}
	\caption{Snapshot of the simulation environment} 
	\label{fig:sim-snap}
\end{figure}

\subsection{Simulation Environment}
To test our method in presence of minimum noise and controllable conditions we first perform the algorithm validation in a simulation environment through ROS \cite{Quigley09} and Gazebo \cite{GazeboKoenig}, as seen in Figure \ref{fig:sim-snap}. To simulate the UAV, UUV and underwater environment RotorS Simulator \cite{Furrer2016} and UUVSim \cite{Manhaes2016} were used. UAV camera was simulated with a standard Gazebo camera plugin, with parameters set to nominal values of our field data acquisition setup.

\subsection{Data Acquisition Setup}

Our data acquisition setup consisted of a \textit{DJI Matrice 210 V2} \cite{Matrice210v2} UAV equipped with a \textit{DJI Zenmuse X5S} gimbaled camera using a \textit{DJI MFT 15mm/1.7 ASPH} lens and BlueROV2 \cite{BlueROV2} UUV.
Parameters of the camera intrinsic matrix and radial-tangential distortion model were calibrated using the \textit{Kalibr} toolbox \cite{Furgale2013}. Experiments were conducted by the UAV surveying the shallow docking area in front of the Laboratory of Mariculture  University of Dubrovnik research facility at Bistrina, Croatia. It is important to note that altitude measurements of the UAV in our setup represent the altitude above the takeoff point. To get the correct distance of the UAV from the water surface we offset the altitude by distance from the takeoff point to current sea level, which we obtained by tape-measuring. Depending on the use case, this can be bypassed by estimating the height of the UAV from the sea level by using known dimensions of a reference object, e.g. the UUV itself at the water surface.

\subsubsection{Ground truth system}
\begin{figure}[!htb]
	\centering
	\includegraphics[width=0.95\columnwidth]{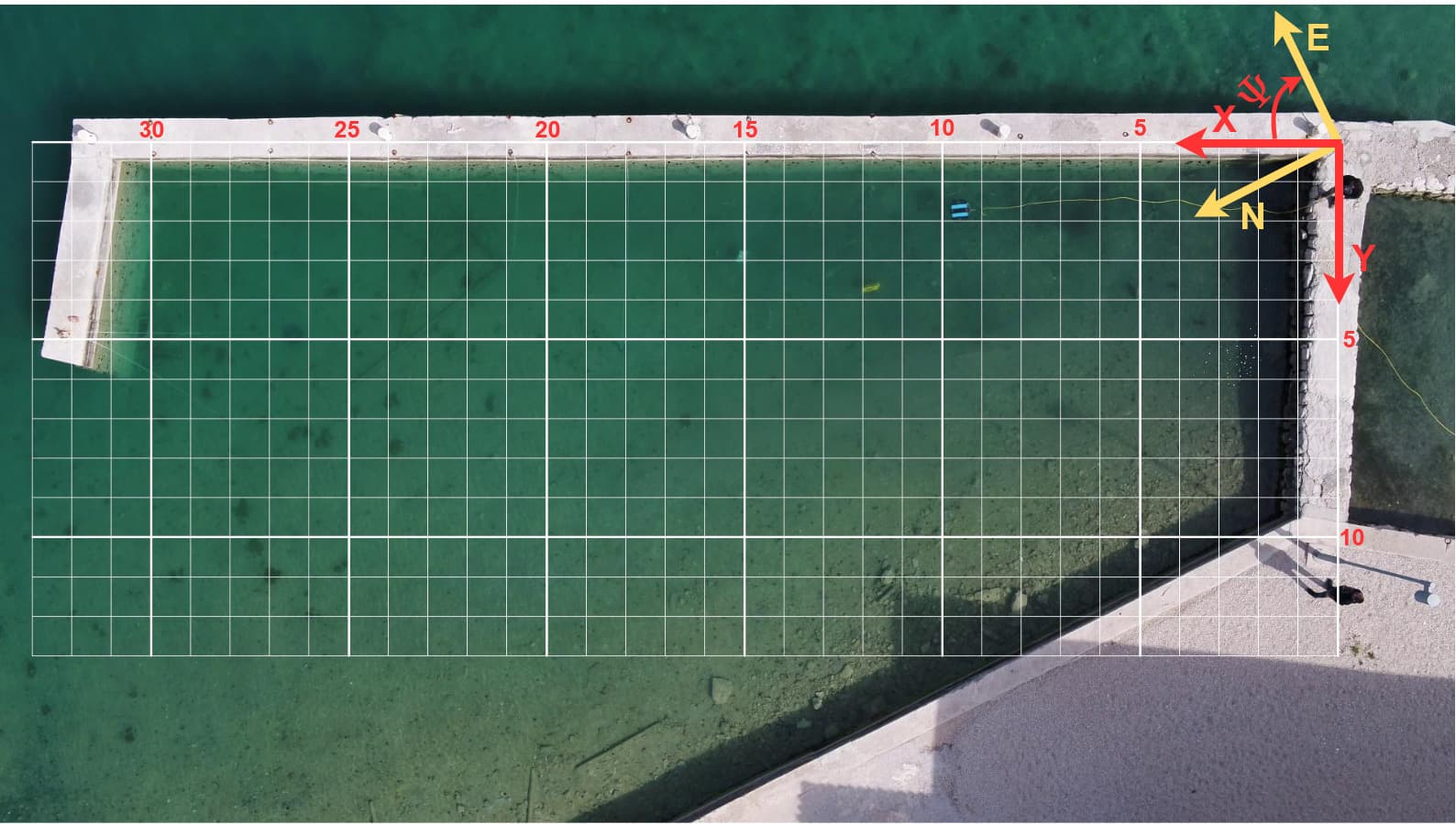}
	\caption{The ground truth system}
	\label{fig:groundtruth}
\end{figure}

\begin{figure}[!htb]
	\centering
	\includegraphics[width=0.8\columnwidth]{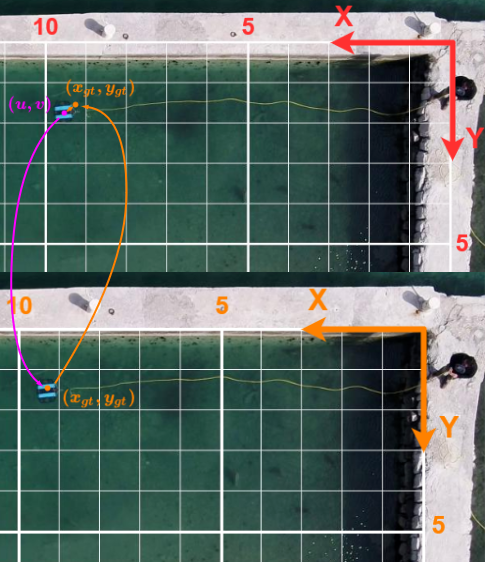}
	\caption{Since cameras project points from 3-D scene to 2-D image plane, meter per pixel can vary on different pixel locations and appropriate scaling has to be performed for locations with  different distances from the camera. To obtain position of the UUV in the ground truth system (top), we read its location in the re-scaled system (bottom).}
	\label{fig:rescale-grid}
\end{figure}

The ground truth coordinate system is a convenience frame used for the evaluation and visualization of recovered positions. The exact coordinates of the system are obtained by tape-measuring the distances between landmarks. The origin of this frame is rotated around the Z-axis relative to the ENU frame by $\Psi = -67.3\degree$, as shown in Figure \ref{fig:groundtruth}. Therefore, 3D position of the UUV in the ground truth coordinate system ${}^{GT}P$ can be obtained by transformation:
\begin{equation}
{}^{GT}P = R_{z}(-67.3\degree) {}^D P + {}^{GT} T \nonumber
\end{equation}
where ${}^{GT} T$ is the translational offset from GT system origin to the UAV, obtained based on grid shown in Figure \ref{fig:groundtruth} .
In all experiments, UAV was hovering with camera gimbaled at an $90\degree$ pitch angle. To compensate for small deviations in UAV positon, we use MIL object tracker to track the origin position in subsequent frames. This enabled us to precisely compute the origin-UUV vector in each frame and made the retrieval of ground truth positions straightforward. For all points on the image with ground truth coordinates $(x,y,0)$, their corresponding x and y components can be computed directly from the grid. However, in all experiments UUV is at some depth $z = -d_{UUV}$ from the origin. To obtain position of the UUV in the ground truth system , we re-scale the initial grid with respect to the depth $d_{UUV}$ and compute its position in the re-scaled grid (Figure \ref{fig:rescale-grid}).

\begin{figure*}[t!]
	\begin{minipage}{.5\linewidth}
		\centering
		\subfloat[]{\label{main:a}\includegraphics[scale=.4]{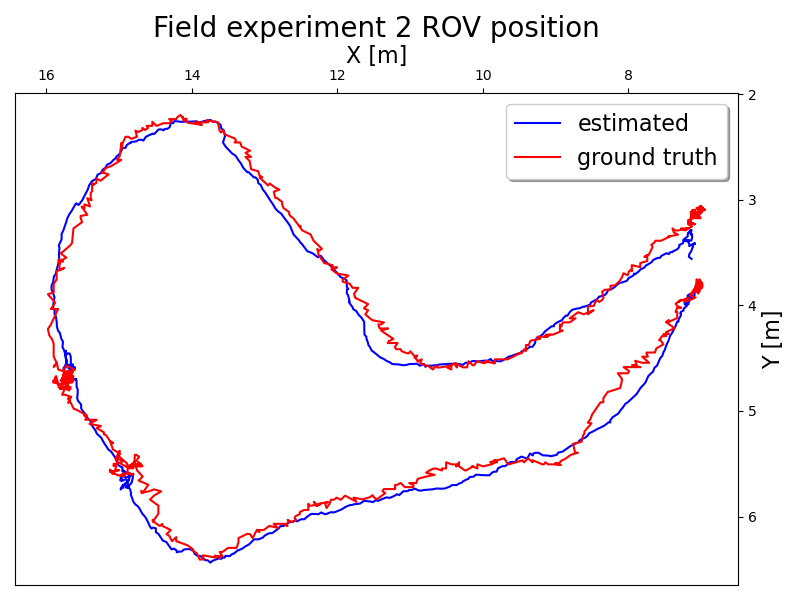}}
	\end{minipage}%
	\begin{minipage}{.5\linewidth}
		\centering
		\subfloat[]{\label{main:b}\includegraphics[scale=.4]{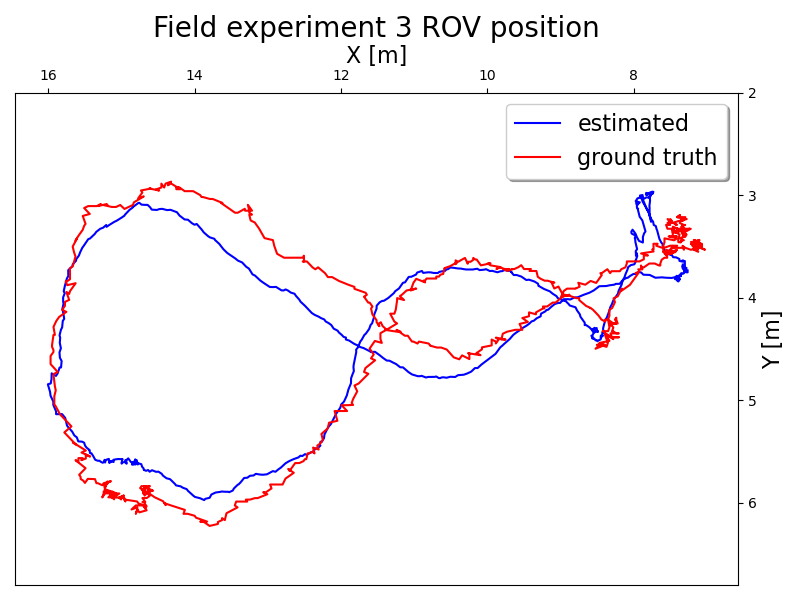}}
	\end{minipage}\par\medskip
	\centering
	\subfloat[]{\label{main:c}\includegraphics[scale=.4]{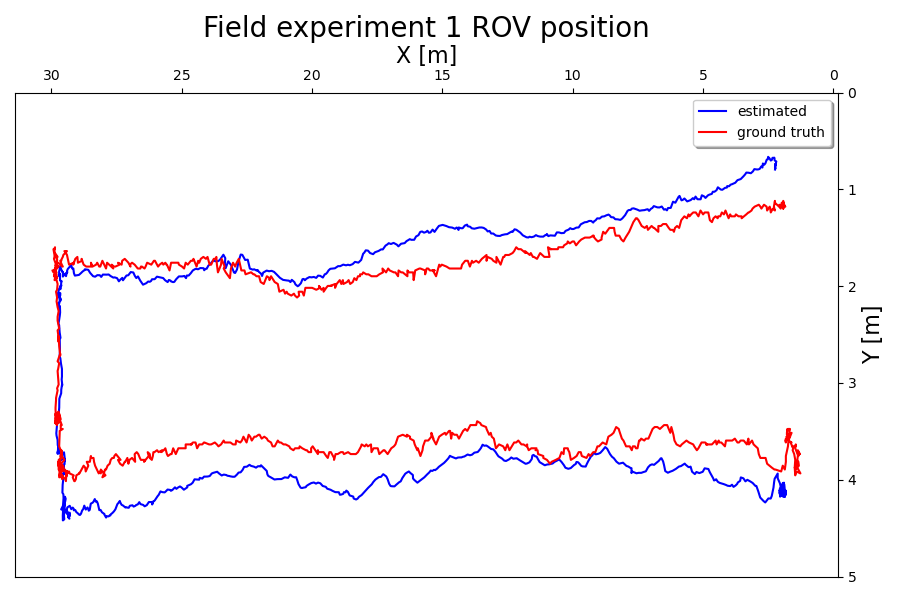}}
	\caption{Recovered $x-y$ positions and ground truth $x-y$ positions for field experiments}
	\label{fig:results-exp}
\end{figure*}
\begin{figure}[!htb]
	\vspace{-0.3cm}
	\centering
	\includegraphics[width=0.85\columnwidth]{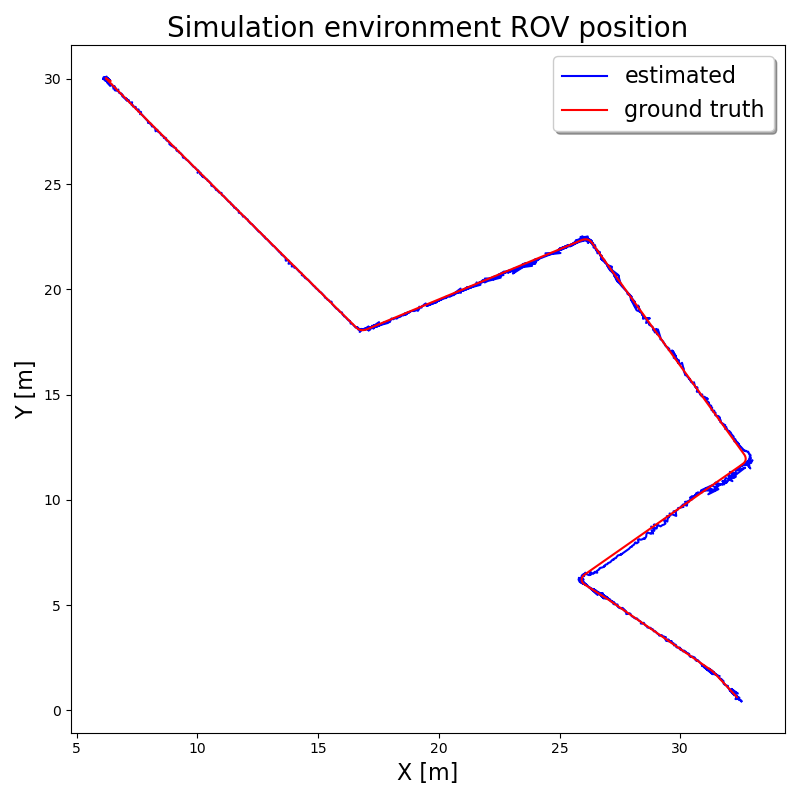}
	\caption{Recovered and ground truth $x-y$ positions of the UUV in simulation environment}
	\label{fig:results-sim}
\end{figure}
\subsection{Discussion of results}
Plot of the recovered $x-y$ positions of the UUV for simulation and field experiments are shown on Figure \ref{fig:results-sim} and Figure \ref{fig:results-exp}. Average errors per experiment are summarized in Table \ref{tab:results}.  As expected the position estimation inside the simulation yields the smallest error. Since all camera parameters (intrinsic and extrinsic) are known, most of the error comes from tracker deviations. Although these deviations are small, due to high altitudes they sum to significant error. In the experiment 1 UUV maintained constant depth of 0.63 meters while in other two experiments depth varied from 0.21 to 1.95 meters. To account for the displacements of the hovering UAV an additional object tracker was employed for tracking the origin image coordinates, which accounts for the noise in the ground truth trajectory. Results obtained from all 3 field experiments have a mean error under 0.5 m indicating that method presented in this paper can be used as reliable alternative to USBL in shallow waters and as a source of ground truth for position estimation algorithms.

\begin{comment}
\begin{figure}[htb!]
	\vspace{-0.3cm}
	\centering
	\includegraphics[width=0.75\columnwidth]{images/results/simulation.png}
	\caption{Recovered and ground truth $x-y$ positions of the UUV in simulation environment}
	\label{fig:results-sim}
\end{figure}
\end{comment}
%%%%%%%%%%% Results table %%%%%%%%%%%%
\bgroup
\def\arraystretch{1.5}%  1 is the default, change whatever you need
\begin{table}[!htb]
	\centering
	\begin{tabular}{ccc}
		\textbf{}             & \textbf{Mean Absolute Error [m]} & \textbf{RMSE [m]} \\ \hline
		\textbf{Simulation}   &                  0.142               &         0.209     \\ \hline
		\textbf{Experiment 1} &                  0.372               &         0.442     \\ \hline
		\textbf{Experiment 2} &                  0.293               &         0.370     \\ \hline
		\textbf{Experiment 3} &                  0.394               &         0.459      \\ \hline
	\end{tabular}
	\caption{Average position errors per experiment}
	\label{tab:results}
\end{table}
\egroup

\section{Conclusion and Further Research}
In this work we present the approach to UUV position estimation from image sequence obtained by the UAV surveying the operating area combined with measurements of altitude and depth from the UAV and UUV, respectively. Through simulation and field experiments we show the presented approach can be utilized as a cost-effective ground truth source for UUV position estimation, dataset collection and obtaining geo-referenced data in shallow-water environments. We provided a quantitative assessment of recovered 3D positions by comparing our results with tape-measured ground truth and showed small error rates. Our further work might focus on addressing the potential weakness of this method, such as increasing the tracking robustness in the presence of occlusions and modeling the distortion introduced by the refraction at the water surface. Additionally, more extensive analysis can be conducted on how choice of the tracking algorithm or camera parameters affects the performance of this estimation method.
%%%%%%%%%%%%%%%%%%%%%%%%%%%%%%%%%%%%%%%%%%%%%%%%%%%%%%%%%%%%%%%%%%%%%%%%%%%%%%%%

\balance
\bibliographystyle{literatura/IEEEtran}
\bibliography{literatura/literatura}

\end{document}